\documentclass[10pt,twocolumn,letterpaper]{article}

\usepackage[applications]{wacv}
\usepackage{xcolor}
\usepackage{colortbl}
\usepackage{multirow}

\usepackage{float}
\usepackage{tabularx} 

\usepackage{graphicx} 
\usepackage{multirow}  
\usepackage{colortbl}
\usepackage{booktabs}
\usepackage{tabularx}
\usepackage{subcaption}
\usepackage{algorithm}
\usepackage{algorithmic}

\definecolor{wacvblue}{rgb}{0.21,0.49,0.74}
\usepackage[pagebackref,breaklinks,colorlinks,allcolors=wacvblue]{hyperref}

\title{AGGRNet: Selective Feature Extraction and Aggregation for Enhanced Medical Image Classification}

\author{
Ansh Makwe$^{1*}$,
Akansh Agrawal$^{1*}$,
Prateek Jain$^{1*}$,
Akshan Agrawal$^{1*}$,
and Priyanka Bagade$^{1\dagger}$\\
$^{1}$Indian Institute of Technology Kanpur, India\\[3pt]
\small\ttfamily
\begin{tabular}{c}
anshmakwe24@iitk.ac.in,\ akanshcs2020@gmail.com,\\
prateekjain856@gmail.com,\ akshancs2020@gmail.com,\\
pbagade@iitk.ac.in
\end{tabular}
}

\begin{document}
\maketitle
\let\thefootnote\relax
\footnotetext{
$^{*}$ The first four co-authors are designated as first authors and contributed equally.   
$^{\dagger}$ Corresponding author.\\
© 2025. The copyright of this document resides with its authors. 
It may be distributed unchanged freely in print or electronic forms.
}




\begin{abstract}

Medical image analysis for complex tasks such as severity grading and disease subtype classification poses significant challenges due to intricate and similar visual patterns among classes, scarcity of labeled data, and variability in expert interpretations. Despite the usefulness of existing attention-based models in capturing complex visual patterns for medical image classification, underlying architectures often face challenges in effectively distinguishing subtle classes since they struggle to capture inter-class similarity and intra-class variability, resulting in incorrect diagnosis. 
To address this, we propose AGGRNet framework to extract informative and non-informative features to effectively understand fine-grained visual patterns and improve classification for complex medical image analysis tasks. Experimental results show that our model achieves state-of-the-art performance on various medical imaging datasets, with the best improvement up to 5\% over SOTA models on the Kvasir dataset. 

\end{abstract}

\section{Introduction}
\label{sec:intro}

Medical image classification plays a crucial role in accurate diagnosis and effective treatment planning by classifying images into disease subtypes and severity levels. However, despite its criticality, the field remains largely subjective, as annotations by experts often vary. For instance, in disease severity scoring for ulcerative colitis (UC), clinicians often rely on the Mayo Endoscopic Score (MES) \cite{sharara2022assessment}, which ranges from 0 (normal or inactive disease) to 3 (severe disease with spontaneous bleeding and large ulcers), yet they frequently disagree on grading due to subtle visual cues, with even the same expert potentially providing inconsistent scores over time \cite{hashash2024inter}. 

Similarly, classification of specific disease subtypes, such as distinguishing among polyps, esophagitis, ulcerative colitis, or other anatomical landmarks in datasets like Kvasir \cite{Kvasir} (which categorizes GI tract images into eight classes, including pathological findings and endoscopic procedures) can be inconsistent, further amplifying the uncertainty in diagnosis.

\begin{figure}[H]
    \centering
    \includegraphics[width=0.485\textwidth]{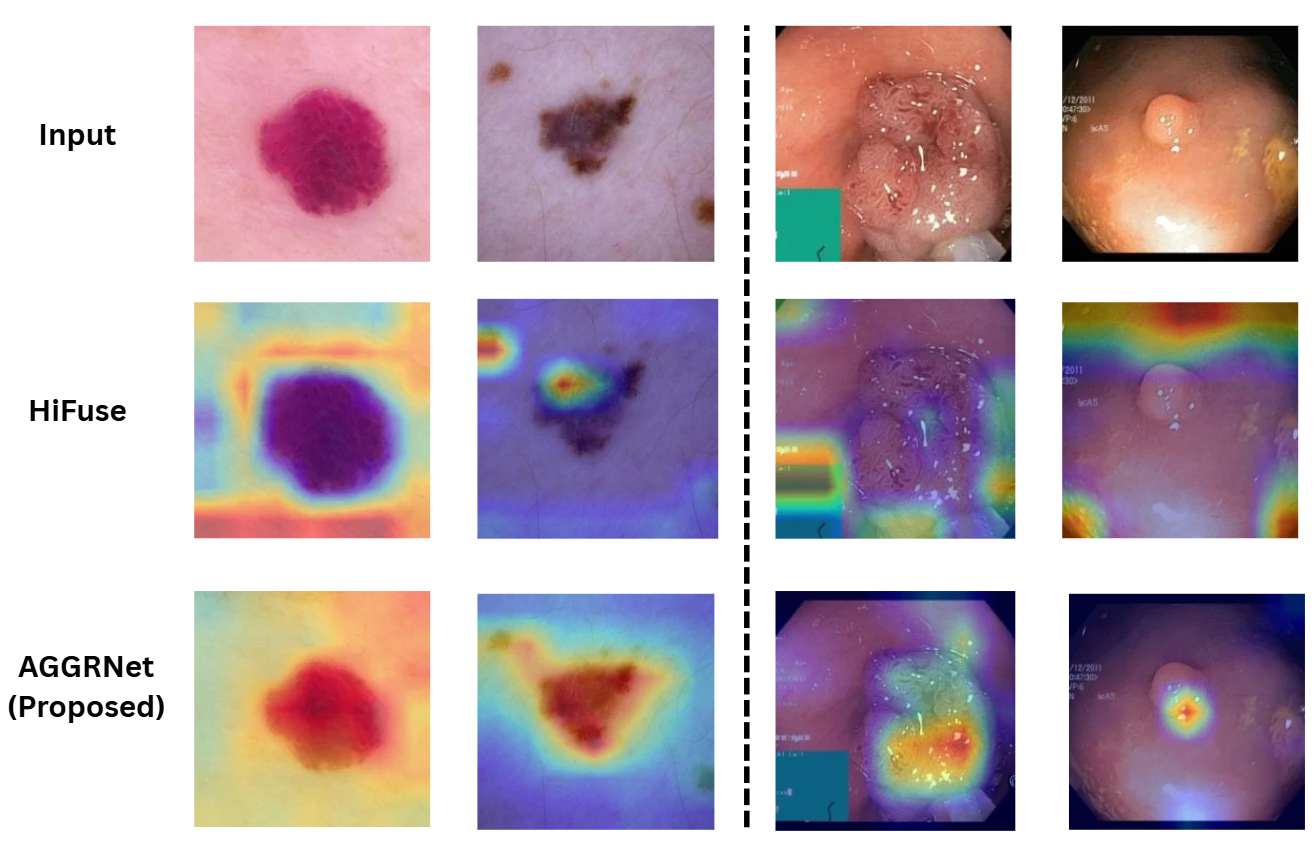} 
    \caption{Comparison of proposed AGGRNet framework with state-of-the-art HiFuse \cite{huo2024hifuse} architecture, showing improved identification of critical regions (Grad-CAM visual results).}
    \label{gradcam}
\end{figure}

\begin{figure*}[h]
    \centering
    \includegraphics[width=0.85\textwidth]{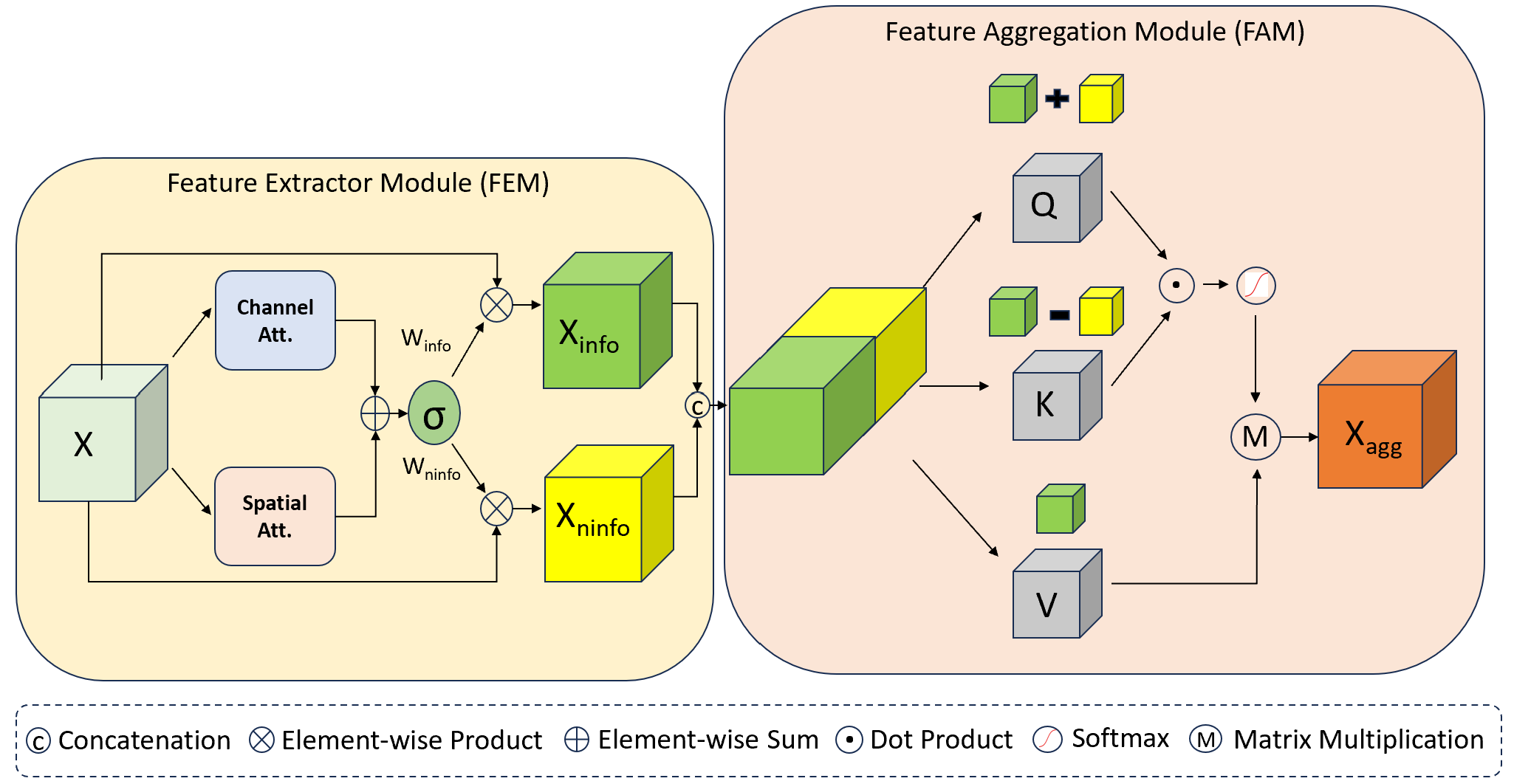} 
    \caption{The proposed Feature Extraction and Aggregation (FEA) Module}
    \label{fea}
\end{figure*}

The inconsistency observed in disease subtype classification and severity grading arises from the overlap of visual patterns among classes. This highlights the inherent similarity between distinct classes as well as the variability of features within the same class, which can potentially result in incorrect diagnosis. 
Although the traditional deep learning approaches \cite{huang2017densely, huo2024hifuse} have demonstrated remarkable success in medical imaging analysis, their feature extraction layers do not capture the inter-class similarity and intra-class variability, making it a challenging task to classify into classes having subtle differences (Figure \ref{gradcam}).

To address this issue, we utilized two key ideas: first, medical images often contain subtle but critical details that require careful separation of relevant features from background features; second, anatomical and pathological patterns often require an understanding of global spatial relationships and contextual information. Building upon that, we propose the AGGRNet framework (Figure \ref{aggrnet}), which incorporates a novel Feature Extraction and Aggregation (FEA) module (Figure \ref{fea}). The FEA module comprises two main components: (1) a Feature Extraction Module (FEM), which we design to separate clinically relevant regions of interest (hereafter referred to as Informative Features) from background (or, Non-Informative Features), and (2) a Feature Aggregation Module (FAM), which captures global dependencies through cross-attention mechanism \cite{lin2022cat}. 

\begin{figure}[h]
    \centering
    \includegraphics[width=0.5\textwidth]{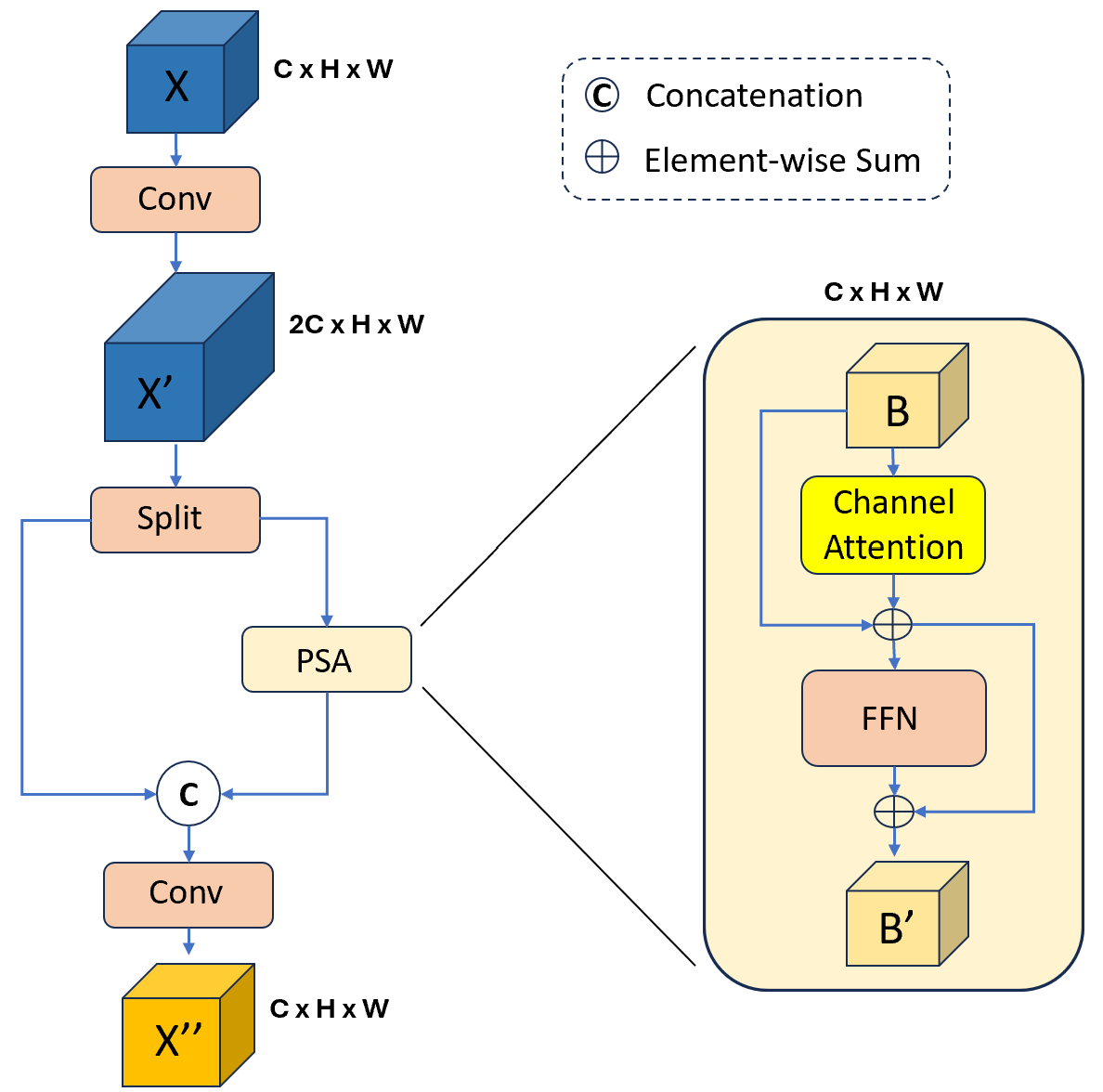} 
    \caption{Cross Stage Partial Channel Attention Block (C2PCA)}
    \label{c2pca}
\end{figure}

The proposed AGGRNet utilizes the YOLOV11 classification backbone \cite{Khanam2024YOLOv11AO} as a base model and incorporates the FEA module for selective feature extraction and aggregation. Although the YOLOv11 is an object detection model, it also includes an efficient and modular classification backbone architecture. With the ablation study in Section \ref{abalation}, we assert that incorporating the FEA module into the YOLOv11 classification backbone enhances the performance of medical image classification. We also replaced the standard C2PSA (cross-stage partial with self-attention) block in YOLOv11 classification backbone with proposed C2PCA (cross-stage partial with channel attention) block (Figure \ref{c2pca}) to further improve performance for two complementary diagnostic objectives: severity grading and disease-subtype classification.

Through extensive experiments, the proposed AGGRNet framework outperforms existing state-of-the-art models. It achieves a performance improvement of more than 5\% for the classification of disease subtypes in the Kvasir \cite{Kvasir} dataset and a performance improvement of more than 2\% for severity classification in the LIMUC dataset \cite{LIMUC}.


\noindent The main contributions can be summarized as follows: 
\begin{itemize}
    \item We propose a novel AGGRNet framework, improving the diagnostic accuracy across disease subtype classification and severity grading.
    \item As a part of this framework, we propose a novel Feature Extraction and Aggregation (FEA) module. It consists of Feature Extraction Module (FEM) and Feature Aggregation Module (FAM), for extracting informative and non-informative features along with global dependencies.
    \item In addition, we propose the C2PCA block to further emphasize the most relevant features. 
    \item AGGRNet framework achieves the best classification performance on five medical image datasets (ordinal and disease-subtype classification datasets), outperforming the SOTA models.

\end{itemize}


\section{Related Works}
\label{sec:formatting}


Deep learning architectures for medical image classification have evolved from CNNs such as VGG \cite{VGG} and ConvNeXt \cite{ConvNext}, DenseNet121 \cite{huang2017densely}, to MLP-based models like MLP-Mixer \cite{mlp-mixer}, and to Transformer-based approaches including ViT \cite{ViT}, DeiT \cite{DeiT-base}, T2T-ViT \cite{T2T-Vit}, and Swin \cite{Swin}, which leverage self-attention for global feature modeling. Transformer models treat images as sequences of patches, enabling the capture of long-range dependencies and global semantic information. However, they often lack local inductive bias and require high computational resources. 



Polat et al. \cite{Polat22} proposed the Class Distance Weighted Cross-Entropy (CDW-CE) loss function for the LIMUC dataset, which addresses ordinal regression challenges in MES scoring by penalizing distant class mispredictions more heavily than conventional cross-entropy loss. While their approach achieved superior performance using ResNet18 \cite{he2016deep}, Inception-v3 \cite{szegedy2015going}, and MobileNet-v3-large \cite{howard2019searching} architectures and demonstrated improved class activation maps, it fundamentally relies on standard feature extraction mechanisms that treat all spatial regions and feature channels uniformly, whereas the severe region is not spread uniformly throughout the colon.



Recent developments in ulcerative colitis severity estimation have explored patient-level multiple instance learning (MIL) approaches to leverage clinical diagnostic records. Shiku et al. \cite{OrdinalMultiInstance} proposed the Selective Aggregated Transformer for Ordinary MIL (SATOMIL) for estimating the severity of ulcerative colitis. Although SATOMIL outperforms conventional MIL methods through selective aggregation with specialized tokens, it operates exclusively within patient-level bag structures requiring multiple images per patient, limiting its applicability to standard single-image classification tasks. 

Moreover, such ordinal loss function-centric or ordinal multiple instance learning approaches remain task-specific to severity grading and do not address the broader challenge of general disease subtype classification across different medical imaging tasks, significantly limiting their clinical applicability and generalizability, particularly when subtle inter-class variations require sophisticated feature discriminative capabilities.

For disease subtype classification, HiFuse \cite{huo2024hifuse} introduced a hierarchical multi-scale feature fusion network to extract spatial context and semantic representations. It has demonstrated strong performance on multiple disease subtype classification datasets. However, it fails to capture the inter-class similarity and intra-class variability, resulting in suboptimal results as shown in Figure \ref{gradcam}. Our proposed AGGRNet model achieved better classification performance while considering global dependencies and effectively segregating relevant and background features.

\section{Methodology}
\label{sec:method}

\begin{figure*}[h]
    \centering
    \includegraphics[width=0.8\textwidth]{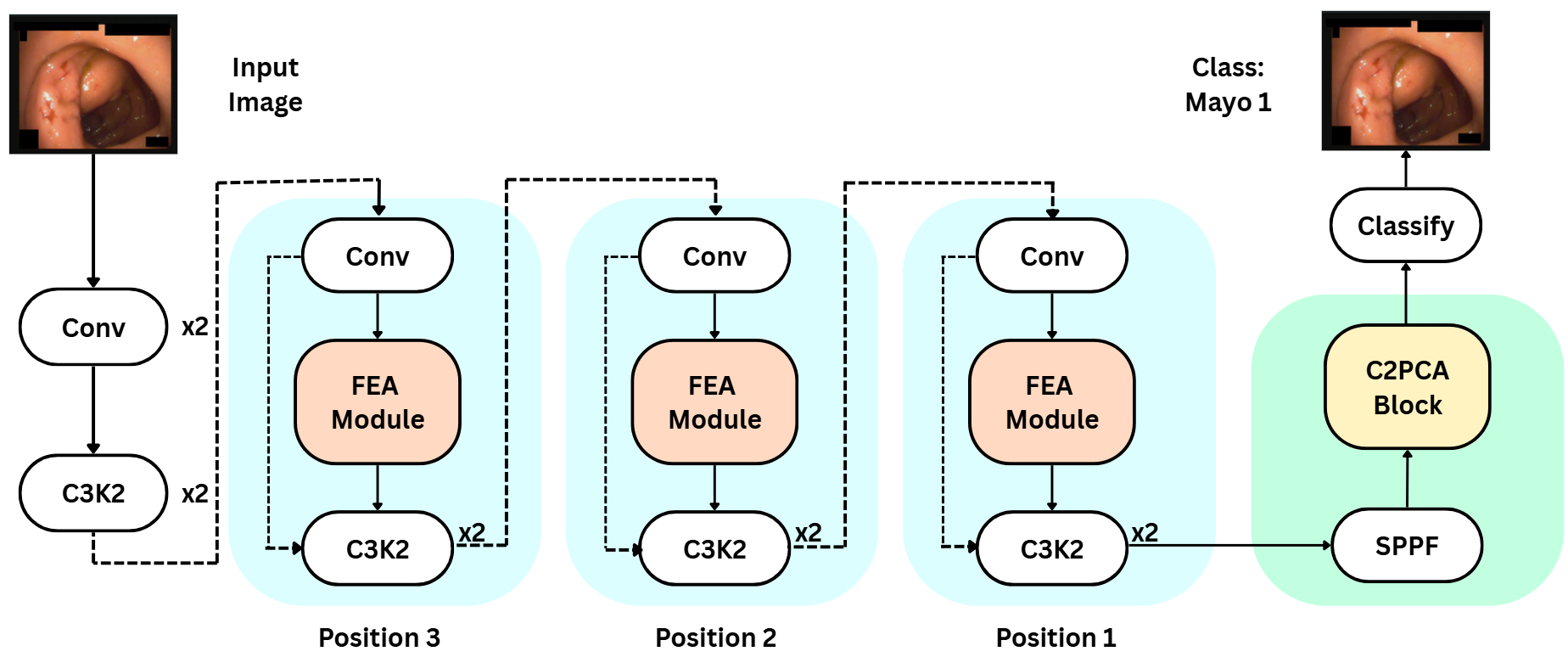} 
    \caption{The proposed architecture AGGRNet with the novel FEA module and C2PCA block.}
    \label{aggrnet}
\end{figure*}

In this section, we describe the proposed AGGRNet framework for medical image classification. Section \ref{feature_extraction_aggregation} explains the novel Feature Extraction and Aggregation Module (FEA). Then, we discuss its integration with CNN architectures and the overall AGGRNet architecture design in Section \ref{cnn} and \ref{proposed_arch} respectively.


\subsection{Feature Extraction and Aggregation Module}
\label{feature_extraction_aggregation}

The proposed FEA module (Figure \ref{fea}) consists of two components: (1) a Feature Extraction Module (FEM) that separates informative and non-informative features from input feature maps, (2) a Feature Aggregation Module (FAM) that captures global-level relationships through cross-attention mechanisms. 

\subsubsection{Feature Extraction Module (FEM)}

The FEM addresses the critical challenge of distinguishing between diagnostically relevant features and background in medical images. Given an input feature map $\mathbf{X} \in \mathbb{R}^{C \times H \times W}$, where $H$, $W$, and $C$ represent height, width, and number of channels respectively. The proposed FEM decomposes $\mathbf{X}$ into two distinct components: informative features $\mathbf{X}_{\text{info}}$ and non-informative features $\mathbf{X}_{\text{ninfo}}$. In order to identify $\mathbf{X}_{\text{info}}$ and $\mathbf{X}_{\text{ninfo}}$,  the FEM module employs spatial and channel attention mechanisms \cite{woo2018cbam} to generate attention weights. The spatial attention module $\text{SA}(\cdot)$ focuses on identifying spatially significant regions, while the channel attention module $\text{CA}(\cdot)$ emphasizes important feature channels:
\begin{align}
\text{SA}_{\text{out}} &= \text{SA}(\mathbf{X}) \label{eq:1} \\
\text{CA}_{\text{out}} &= \text{CA}(\mathbf{X}) \label{eq:2}
\end{align}

The outputs from both attention modules are aggregated with element-wise addition and passed through a sigmoid activation function to generate normalized attention scores:
\begin{equation}
\mathbf{S} = \sigma(\text{SA}_{\text{out}} + \text{CA}_{\text{out}}) \label{eq:3}
\end{equation}

where $\sigma$ denotes the sigmoid function, ensuring $\mathbf{S} \in \mathbb{R}^{C \times H \times W}$.

A key idea of our approach is the introduction of ``Adaptive Thresholding for Feature Segregation'', a learnable threshold parameter $\tau$, which adaptively determines the boundary between informative and non-informative features through the training. This threshold serves as a decision boundary that classifies each element in the attention score matrix $\mathbf{S}$ as either contributing to informative or non-informative features. The initial value of $\tau$ is set to 0.5, providing equal weightage to the identification of both informative and non-informative features at the beginning of the training process. The thresholding operation generates two complementary binary weight matrices $\mathbf{W}_{\text{info}}$ and $\mathbf{W}_{\text{ninfo}}$, which are constructed using the rule:
\begin{equation}
\mathbf{W}_{\text{info}}[i,j,k] = \begin{cases} 
1 & \text{if } \mathbf{S}[i,j,k] \geq \tau \\
0 & \text{otherwise}
\end{cases} \label{eq:4}
\end{equation}

\begin{equation}
\mathbf{W}_{\text{ninfo}}[i,j,k] = \begin{cases} 
1 & \text{if } \mathbf{S}[i,j,k] < \tau \\
0 & \text{otherwise}
\end{cases} \label{eq:5}
\end{equation}

where $\mathbf{S}[i,j,k]$ represents the attention score at spatial location $(i,j)$ and channel $k$. These binary matrices act as selective masks, with $\mathbf{W}_{\text{info}}$ identifying regions and channels deemed informative (attention scores above threshold), while $\mathbf{W}_{\text{ninfo}}$ captures the complementary non-informative components. Using these binary matrices, the segregated feature representations are obtained through element-wise multiplication as:
\begin{align}
\mathbf{X}_{\text{info}} &= \mathbf{W}_{\text{info}} \odot \mathbf{X} \label{eq:6} \\
\mathbf{X}_{\text{ninfo}} &= \mathbf{W}_{\text{ninfo}} \odot \mathbf{X} \label{eq:7}
\end{align}

where $\odot$ denotes element-wise multiplication.

The FEM module incorporates three sets of learnable parameters: (1) spatial attention weights that capture spatial dependencies, (2) channel attention weights that model inter-channel relationships, and (3) the adaptive threshold $\tau$ that evolves during training to optimize the separation boundary. This design enables the module to automatically learn the optimal distinction between informative and non-informative features throughout the training.

\subsubsection{Feature Aggregation Module (FAM)}

The FAM leverages cross-attention mechanisms to enable informative features to attend to non-informative features, boosting the model's capability to capture global-level contextual relationships. It employs a novel cross-attention formulation where the Query (Q), Key (K), and Value (V) are defined to maximize the focus on the most informative features as:
\begin{align}
\mathbf{Q} &= \mathbf{X}_{\text{info}} + \mathbf{X}_{\text{ninfo}} \label{eq:8} \\
\mathbf{K} &= \mathbf{X}_{\text{info}} - \mathbf{X}_{\text{ninfo}} \label{eq:9} \\
\mathbf{V} &= \mathbf{X}_{\text{info}} \label{eq:10}
\end{align}


\subsubsection{Mathematical Justification for Q-K Formulation}
The mathematical formulation of $\mathbf{Q}$ and $\mathbf{K}$ is designed to create a bias toward informative features $\mathbf{X}_{\text{info}}$. When computing attention weights through $\mathbf{Q}\mathbf{K}^T$, our formulation yields:
\begin{align}
\mathbf{Q}\mathbf{K}^T &= (\mathbf{X}_{\text{info}} + \mathbf{X}_{\text{ninfo}}) (\mathbf{X}_{\text{info}} - \mathbf{X}_{\text{ninfo}})^T \nonumber \\
&= \mathbf{X}_{\text{info}}\mathbf{X}_{\text{info}}^T - \mathbf{X}_{\text{info}}\mathbf{X}_{\text{ninfo}}^T \nonumber \\
&\quad + \mathbf{X}_{\text{ninfo}}\mathbf{X}_{\text{info}}^T - \mathbf{X}_{\text{ninfo}}\mathbf{X}_{\text{ninfo}}^T \nonumber \\
&= \|\mathbf{X}_{\text{info}}\|^2 - \|\mathbf{X}_{\text{ninfo}}\|^2 \nonumber \\
&\quad + (\mathbf{X}_{\text{ninfo}}\mathbf{X}_{\text{info}}^T - \mathbf{X}_{\text{info}}\mathbf{X}_{\text{ninfo}}^T) \label{eq:qk_components}
\end{align}

This expansion in Equation \ref{eq:qk_components} reveals three key mathematical components that create the informative bias:

\begin{itemize}
\item \textbf{Positive Bias Term for Informative Regions ($\|\mathbf{X}_{\text{info}}\|^2$)}: Creates high attention scores at spatial locations where informative features have large magnitudes, directly promoting regions rich in discriminative information.

\item \textbf{Negative Bias Term against Non-Informative Regions ($-\|\mathbf{X}_{\text{ninfo}}\|^2$)}: Suppresses attention scores where non-informative features dominate, effectively filtering out irrelevant spatial regions.

\item \textbf{Cross-Modal Interaction Terms}: The $(\mathbf{X}_{\text{ninfo}}\mathbf{X}_{\text{info}}^T - \mathbf{X}_{\text{info}}\mathbf{X}_{\text{ninfo}}^T)$ terms capture complex relationships between informative and non-informative components, enabling subtle feature interactions.
\end{itemize}

When $\|\mathbf{X}_{\text{info}}\| >> \|\mathbf{X}_{\text{ninfo}}\|$ at a spatial location, $\mathbf{Q}\mathbf{K}^T$ produces high attention weights because the positive $\|\mathbf{X}_{\text{info}}\|^2$ term dominates while the negative $\|\mathbf{X}_{\text{ninfo}}\|^2$ term remains small. Thus, the formulation will promote learning of informative features during backpropagation, as gradients would flow preferentially through high-attention regions. Therefore, this mathematical formulation creates a \textbf{contrast-based cross attention mechanism} where:

\begin{itemize}
\item $\mathbf{Q} = \mathbf{X}_{\text{info}} + \mathbf{X}_{\text{ninfo}}$ represents the complete feature context
\item $\mathbf{K} = \mathbf{X}_{\text{info}} - \mathbf{X}_{\text{ninfo}}$ acts as a discriminative filter encoding informativeness
\item $\mathbf{V} = \mathbf{X}_{\text{info}}$ ensures only informative features are aggregated in the final output
\end{itemize}

This formulation aligns with attention mechanism principles where similarity between queries and keys determines relevance, with the $\mathbf{K}$ formulation encoding the degree of informativeness at each spatial location.
\begin{equation}
\text{Attention}(\mathbf{Q},\mathbf{K},\mathbf{V}) = \text{softmax}\left(\frac{\mathbf{Q}\mathbf{K}^T}{\sqrt{d_k}}\right)\mathbf{V} \label{eq:11}
\end{equation}
where $d_k$ represents the dimension of the key vector, and the softmax function ensures normalization of attention weights. The final aggregated features $\mathbf{X}_{\text{agg}}$ incorporates global contextual information:
\begin{equation}
\mathbf{X}_{\text{agg}} = \text{Attention}(\mathbf{Q},\mathbf{K},\mathbf{V}) \label{eq:12}
\end{equation}

\subsection{Integration with CNN Architectures}
\label{cnn}
The FEA module is designed to be architecture-agnostic and can be seamlessly integrated into existing CNN frameworks. Based on our analysis of CNN feature hierarchies, we propose positioning the FEA module between the deeper layers of the network, where well-extracted features are available for global relationship modeling, while allowing subsequent convolutional layers to capture local patterns. To preserve important feature information and facilitate gradient flow, we incorporate a skip connection between the input to the FEA module and its output:
\begin{equation}
\mathbf{X}_{\text{output}} = \mathbf{X}_{\text{agg}} + \mathbf{X} \label{residual}
\end{equation}

This residual design in Equation \ref{residual} ensures that the original feature information is retained while augmenting it with globally-aware representations.

\subsection{Proposed Architecture - AGGRNet}
\label{proposed_arch}

Figure \ref {aggrnet} shows the proposed AGGRNet architecture block diagram. We integrate the novel FEA module and the C2PCA module into the classification backbone of the YOLOv11 model. The reason behind choosing the classification backbone of YOLOv11 is to leverage the specially designed C3K2, C2PSA, and SPPF blocks for efficient feature extraction and processing. The C3K2 block is a lightweight cross-stage partial module that enhances feature flow while reducing computational overhead through smaller kernel size convolutions. The C2PSA (Cross Stage Partial with self-attention) module integrates self-attention mechanisms to focus on relevant image regions. Finally, the SPPF (Spatial Pyramid Pooling-Fast) block performs multiple pooling operations to aggregate information from feature maps of various aspect ratios, enabling robust handling of objects at different scales. Together, these components form a streamlined, optimized backbone for image classification tasks. We also propose strategic modifications to the C2PSA block aimed at optimizing performance for medical image classification tasks in Section \ref{c2psa_modified}.

\begin{figure*}[h]
    \centering
    \includegraphics[width=0.85\textwidth]{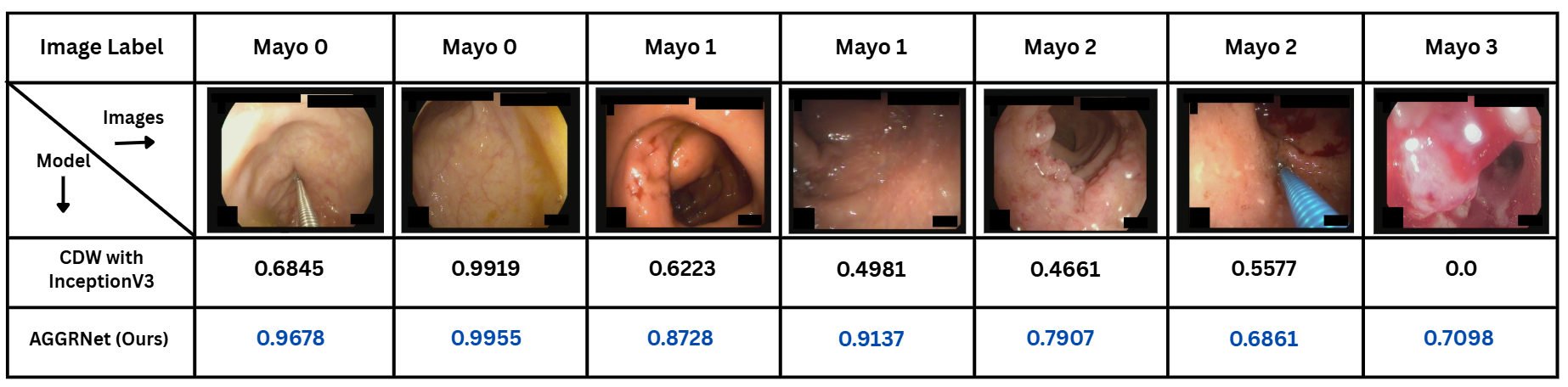} 
    \caption{Class-wise Confidence Score (Predicted Probability) Comparison on LIMUC dataset between state-of-the-art CDW-CE (Inception V3) model and the proposed AGGRNet Framework}
    \label{limuc}
\end{figure*}
\begin{figure*}[h]
    \centering
    \includegraphics[width=0.85\textwidth]{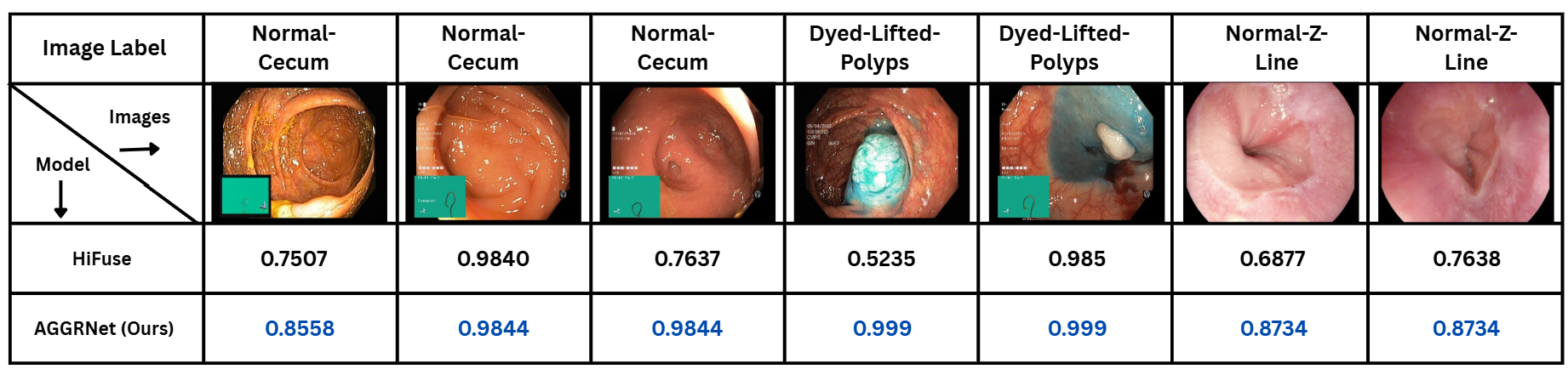} 
    \caption{Class-wise Confidence Score (Predicted Probability) Comparison on Kvasir dataset between state-of-the-art HiFuse model and the proposed AGGRNet Framework}
    \label{kavasir}
\end{figure*}

\subsubsection{C2PSA Module Enhancement}
\label{c2psa_modified}


The original C2PSA module in YOLOv11 utilizes self-attention mechanisms. However, given that our FEA module already extracts globally attended informative features, we replace the self-attention component with a channel attention module, creating the C2PCA (Cross Stage Partial with Channel Attention) block (Figure \ref{c2pca}). This modification is motivated by the following considerations:
\begin{itemize}
\item \textbf{Feature Complementarity}: Since the FEA module handles spatial and cross-feature attention, the C2PSA module can focus on channel-wise feature prioritization.
\item \textbf{Computational Efficiency}: Channel attention is computationally more efficient than self-attention for the final-layer processing.
\item \textbf{Medical Domain Specificity}: Channel attention helps prioritize different anatomical or pathological patterns encoded in different feature channels.
\end{itemize}

\noindent\textbf{Mathematical Formulation:}
\noindent Given an input feature map $X \in \mathbb{R}^{C \times H \times W}$, the mathematical formulation of the C2PCA module is presented in Algorithm \ref{mathemetical_c2pca}. The C2PCA block first doubles the channel dimension with a $1\times1$ convolution and splits the result into two equal parallel branches. The left branch ($X_A$) is kept intact to preserve original low-level features, while the right branch ($X_B$) is refined by channel-attention weighting and a feed-forward network, each enclosed by residual connections for stability. It enables the module to selectively emphasize the most relevant channels for medical image analysis tasks. Finally, the untouched $X_A$ and the attention-enhanced $X_B'$ are concatenated to get the final feature map (Figure \ref{c2pca}).






\begin{algorithm}[H]
\caption{Mathematical Formulation of C2PCA Module}
\label{mathemetical_c2pca}
\begin{algorithmic}[1]
\REQUIRE Input feature map $X \in \mathbb{R}^{C \times H \times W}$
\STATE $X' \gets \text{Conv}(X)$ \COMMENT{Channel Expansion, $X' \in \mathbb{R}^{2C \times H \times W}$}
\STATE $X_A, X_B \gets \text{Split}(X')$ \COMMENT{Split into two branches, each in $\mathbb{R}^{C \times H \times W}$}
\STATE $X_B^{att} \gets \text{ChannelAttention}(X_B) \odot X_B$ \COMMENT{Apply Channel Attention}
\STATE $X_B^{res1} \gets X_B + X_B^{att}$ \COMMENT{First Residual Connection}
\STATE $X_B^{ffn} \gets \text{FFN}(X_B^{res1})$ \COMMENT{Feed-Forward Network}
\STATE $X_B' \gets X_B^{res1} + X_B^{ffn}$ \COMMENT{Second Residual Connection}
\STATE $X'' \gets \text{Concat}(X_A, X_B')$ \COMMENT{Concatenate branches, final output in $\mathbb{R}^{2C \times H \times W}$}
\RETURN $X''$
\end{algorithmic}
\end{algorithm}

\section{Experiments}
\label{sec:experiments}

This section details the public benchmark datasets employed for evaluating AGGRNet (Section \ref{datasets}), followed by implementation details of the proposed framework (Section \ref{imp_details}). The performance evaluation metrics are defined in Section \ref{metrices}. Comprehensive quantitative results demonstrating the efficacy of AGGRNet are presented in Section \ref{results}, with an ablation study provided in Section \ref{abalation}. Refer to the supplement for additional experiments and results. 


\subsection{Datasets}
\label{datasets}
We evaluated the proposed AGGRNet framework on various publicly available datasets for severity grading (ordinal) and disease subtype classifications. 

\begin{itemize}
\item \textbf{LIMUC \cite{LIMUC}:}
The LIMUC (Labelled Images for Ulcerative Colitis) dataset comprises 11,276 endoscopic images from 564 patients over 1,043 colonoscopy sessions. Each image is annotated with a severity grade based on the Mayo Endoscopic Score (MES), distributed across four classes: MES 0 (6,105 images), MES 1 (3,052 images), MES 2 (1,254 images), and MES 3 (865 images). 

\item \textbf{ISIC 2018 \cite{ISIC1}\cite{ISIC2}:}
The ISIC 2018 Task 3 dataset is part of the challenge of the International Skin Imaging Collaboration (ISIC) and focuses on the classification of skin lesions. It consists of 10,015 dermoscopic images annotated with one of seven diagnostic categories: melanoma (1113), melanocytic nevus (6705), basal cell carcinoma (514), actinic keratosis (327), benign keratosis (1099), dermatofibroma (115), and vascular lesion (142). 

\item \textbf{Kvasir \cite{Kvasir}:}
The dataset contains endoscopic images captured from the gastrointestinal tract. The dataset is divided into eight categories (each of 500 images) - dyed-lifted-polyps, dyed-resection-margins, esophagitis, normal-cecum, normal-pylorus, normal-z-line, polyps, and ulcerative-colitis. 

\item \textbf{PathMNIST \cite{Medmnist}:}
PathMNIST is a collection of 107,180 histopathological images, split into nine classes - adipose (12,784), background (12,580), debris (13,832), lymphocytes (12,520), mucus (14,015), smooth muscle (14,654), normal colon mucosa (10,510), cancer-associated stroma (10,126), colorectal adenocarcinoma epithelium (7,159). 

\item \textbf{RetinaMNIST \cite{Medmnist}:}
RetinaMNIST is a collection of 1,600 retinal images, split into 5 classes indicating diabetic retinopathy severity, distributed as follows: class 0 (714 images), class 1 (186 images), class 2 (326 images), class 3 (282 images), and class 4 (92 images).

\end{itemize}

\subsection{Implementation Details}
\label{imp_details}
The proposed AGGRNet framework was implemented using the PyTorch library and trained on NVIDIA RTX A6000 GPU. The classification backbone is initialized with ImageNet pretrained weights. The model was trained on input images resized to $224\times224$, using stochastic gradient descent (SGD) with an initial learning rate of 0.01, momentum of 0.937, and weight decay of $5\times10^{-4}$.

\subsection{Evaluation Metrics}
\label{metrices}
Model performance was evaluated using Accuracy, Quadratic Weighted Kappa (QWK), Mean Absolute Error (MAE), Precision (P), Recall (R),  F1-score, and Area under the Curve (AUC), capturing different aspects of classification efficacy. 
QWK and MAE are used for regression-like ordinal classification tasks. Metric selection is guided by results reported in the literature on SOTA models for different datasets.

\subsection{Results}
\label{results}

\begin{itemize}
    \item \textbf{Results on LIMUC dataset:} Table \ref{table_all_results_a} shows substantial improvements over the SOTA model (CDW-CE with Inception-v3), including a 2.2\% gain in accuracy, 1.3\% in Macro-f1, 1.3\% in QWK, and a reduction of 0.021 in MAE. Notably, AGGRNet outperforms class-distance-based-cross-entropy approaches by effectively leveraging its feature extraction and aggregation mechanisms to better distinguish between adjacent severity classes (MES 0-3), as evidenced by the superior QWK score. Figure \ref{limuc} visually illustrates the per-class confidence scores returned by both CDW-CE (Inception-v3) and our AGGRNet for representative images from each Mayo class. AGGRNet consistently assigns substantially higher predicted probabilities to the correct classes compared to the baseline.
    \item \textbf{Results on ISIC2018 dataset:} Table \ref{table_isic_kvasir} shows substantial improvements over the strongest baseline (HiFuse-Base), including a 1.25\% gain in accuracy, 1.78\% in Macro-f1, and 5.73\% in precision. Notably, AGGRNet outperforms Transformer-based models, Swin-B and Conformer as evidenced by enhanced precision in distinguishing challenging classes such as melanoma and melanocytic nevus.
    \item \textbf{Results on Kvasir dataset:} Table \ref{table_isic_kvasir} shows substantial improvements over SOTA model (HiFuse-Small), including a 5.48\% gain in accuracy, 5.47\% in Macro-f1, 5.75\% in precision, and 5.47\% in recall. Notably, AGGRNet outperforms hybrid CNN-Transformer models like Conformer and BiFormer as evidenced by enhanced recall in distinguishing categories such as polyps and esophagitis. Figure \ref{kavasir} further illustrates the class-wise confidence of our AGGRNet compared to HiFuse for representative Kvasir images. AGGRNet consistently delivers higher predicted probabilities for the correct classes, even on challenging or subtle cases.
    \item \textbf{Results on PathMNIST and RetinaMNIST datasets:} Table \ref{table_all_results_b} presents the performance comparison of our AGGRNet framework against several baseline methods on the PathMNIST \cite{Medmnist} and RetinaMNIST \cite{Medmnist} datasets from MedMNIST. The results show substantial improvements over the SOTA model (ResNet-50 at 28 resolution), including a 0.6\% gain in AUC and 1.5\% in accuracy for PathMNIST. For RetinaMNIST, AGGRNet outperforms the best baseline (Google AutoML Vision) by 0.6\% in accuracy.
\end{itemize}

\begin{table*}[htbp]
    \centering
    \caption{Performance comparison across LIMUC, PathMNIST, and RetinaMNIST datasets.}
    \label{table_all_results}
    \begin{subtable}{0.48\textwidth}
        \centering
        \caption{LIMUC Dataset}
        \label{table_all_results_a}
        \resizebox{\textwidth}{1.8cm}{
        \begin{tabular}{l||cccc}
            \hline
            \multirow{2}{*}{Methods} & \multicolumn{4}{c}{LIMUC} \\ 
            & Accuracy & Macro-f1 & QWK & MAE \\ 
            \hline\hline
            Cross-entropy with Inception-v3 \cite{Polat22} & 0.760 & 0.683 & 0.836 & 0.253 \\
            CORN with Inception-v3 \cite{shi2023deep}  & 0.760 & 0.683 & 0.843 & 0.250 \\
            CO2 with Inception-v3 \cite{albuquerque2021ordinal} & 0.765 & 0.685 & 0.848 & 0.240 \\
            HO2 with Inception-v3 \cite{albuquerque2021ordinal} & 0.766 & 0.690 & 0.846 & 0.242 \\
            CDW-CE with Inception-v3 \cite{Polat22} & 0.788 & 0.726 & 0.868 & 0.215 \\
            SATOMIL \cite{OrdinalMultiInstance} & 0.690 & 0.674 & 0.826 & - \\
            \rowcolor[gray]{0.9}\textbf{Ours} & \textbf{0.810} & \textbf{0.739} & \textbf{0.881} & \textbf{0.194} \\     
            \hline
        \end{tabular}}
    \end{subtable}\hfill
    \begin{subtable}{0.48\textwidth}
        \centering
        \caption{PathMNIST and RetinaMNIST Datasets}
        \label{table_all_results_b}
        \resizebox{\textwidth}{!}{
        \begin{tabular}{l||cc||cc}
            \hline
            \multirow{2}{*}{Methods} & \multicolumn{2}{c||}{PathMNIST} & \multicolumn{2}{c}{RetinaMNIST} \\
            & AUC & Accuracy & AUC & Accuracy \\
            \hline\hline
            ResNet-18 (28) \cite{deep-residual-learning} & 0.983 & 0.907 & 0.717 & 0.524 \\
            ResNet-18 (224) \cite{deep-residual-learning} & 0.989 & 0.909 & 0.710 & 0.493 \\
            ResNet-50 (28) \cite{deep-residual-learning} & 0.990 & 0.911 & 0.726 & 0.528 \\
            ResNet-50 (224) \cite{deep-residual-learning} & 0.989 & 0.892 & 0.716 & 0.511 \\
            auto-sklearn \cite{Medmnist} & 0.934 & 0.716 & 0.690 & 0.515 \\
            AutoKeras \cite{Medmnist} & 0.959 & 0.834 & 0.719 & 0.503 \\
            Google AutoML Vision \cite{Medmnist} & 0.944 & 0.728 & 0.750 & 0.531 \\
            \rowcolor[gray]{0.9}\textbf{Ours} & \textbf{0.996} & \textbf{0.926} & \textbf{0.745} & \textbf{0.537} \\
            \hline
        \end{tabular}}
    \end{subtable}
\end{table*}

\begin{table*}[h]
    \centering
    \caption{Performance comparison on ISIC2018 and Kvasir datasets.}

    \label{table_isic_kvasir}
    \small
    \renewcommand{\arraystretch}{1.2}
    \setlength{\tabcolsep}{4pt} 
    \begin{tabular}{l|ccccc|cccc}
        \hline
        \multirow{2}{*}{Methods} & \multicolumn{5}{c|}{ISIC2018} & \multicolumn{4}{c}{Kvasir} \\
        & Params(M) & Acc & Macro-F1 & Prec & Rec & Acc & Macro-F1 & Prec & Rec \\
        \hline\hline
        VGG-19 \cite{VGG} & 143.68 & 0.7925 & 0.6183 & 0.6371 & 0.6089 & 0.7775 & 0.7775 & 0.7786 & 0.7783 \\
        Mixer-L/16 \cite{mlp-mixer} & 208.20 & 0.7892 & 0.5988 & 0.6136 & 0.5916 & 0.7430 & 0.7414 & 0.7443 & 0.7434 \\
        T2T-ViT\_t-24 \cite{T2T-Vit} & 64.00 & 0.7759 & 0.5721 & 0.5960 & 0.5594 & 0.7690 & 0.7678 & 0.7760 & 0.7691 \\
        DeiT-base \cite{DeiT-base} & 86.57 & 0.7231 & 0.4101 & 0.4719 & 0.4409 & 0.5215 & 0.4848 & 0.5672 & 0.5229 \\
        ViT-B/16 \cite{ViT} & 86.86 & 0.7832 & 0.6093 & 0.6416 & 0.6052 & 0.7610 & 0.7594 & 0.7649 & 0.7623 \\
        ViT-B/32 \cite{ViT} & 88.30 & 0.7792 & 0.5752 & 0.5874 & 0.5690 & 0.7380 & 0.7350 & 0.7424 & 0.7372 \\
        Swin-B \cite{Swin} & 87.77 & 0.7979 & 0.6395 & 0.6509 & 0.6365 & 0.7730 & 0.7729 & 0.7774 & 0.7744 \\
        Conformer-base-p16 \cite{Conformer} & 83.29 & 0.8266 & 0.7244 & 0.7331 & 0.7166 & 0.8425 & 0.8427 & 0.8445 & 0.8437 \\
        ConvNeXt-B \cite{ConvNext} & 88.59 & 0.7995 & 0.6324 & 0.6490 & 0.6206 & 0.7462 & 0.7441 & 0.7569 & 0.7462 \\
        PerViT-M \cite{PerViT} & 43.04 & 0.8164 & 0.6766 & 0.6819 & 0.6729 & 0.8240 & 0.8230 & 0.8288 & 0.8240 \\
        Focal-B \cite{yang2022focal} & 87.10 & 0.7964 & 0.6288 & 0.6573 & 0.6068 & 0.7800 & 0.7793 & 0.7819 & 0.7801 \\
        UniFormer-B \cite{Uniformer} & 50.02 & 0.8244 & 0.6841 & 0.7067 & 0.6654 & 0.8310 & 0.8304 & 0.8309 & 0.8310 \\
        BiFormer-B \cite{biformer} & 56.04 & 0.8266 & 0.6895 & 0.7266 & 0.6647 & 0.8425 & 0.8426 & 0.8467 & 0.8425 \\
        HiFuse-Tiny \cite{HiFuse} & 82.49 & 0.8299 & 0.7299 & 0.7367 & 0.7287 & 0.8485 & 0.8489 & 0.8496 & 0.8490 \\
        HiFuse-Small \cite{HiFuse} & 93.82 & 0.8359 & 0.7270 & 0.7270 & 0.7314 & 0.8612 & 0.8613 & 0.8625 & 0.8613 \\
        HiFuse-Base \cite{HiFuse} & 127.80 & 0.8585 & 0.7532 & 0.7457 & 0.7658 & 0.8597 & 0.8607 & 0.8629 & 0.8601 \\
        \rowcolor[gray]{0.9}\textbf{Ours} & \textbf{38.65} & \textbf{0.871} & \textbf{0.771} & \textbf{0.803} & \textbf{0.748} & \textbf{0.916} & \textbf{0.916} & \textbf{0.920} & \textbf{0.916} \\
        \hline
    \end{tabular}
\end{table*}

\subsection{Ablation Study}
\label{abalation}

To systematically assess the contribution of each architectural component in AGGRNet, we perform a detailed ablation study inspired by the methodology of progressive architectural modification.

\subsubsection{Effect of C2PCA Block}
We start by evaluating a plain YOLOv11 classification backbone with the C2PSA module, and then replace C2PSA with C2PCA to study the effect of our improved attention mechanism. Table \ref{c2pca_abalation} shows that replacing C2PSA with the modified C2PCA block increase the accuracy by +2\%.
\begin{table}[H]
    \centering
    \caption{Effect of adding C2PCA block.}  
    \label{c2pca_abalation}
    \small
    \begin{tabularx}{\linewidth}{l|c}
        \hline
        \textbf{Architecture} & \textbf{Accuracy} \\ 
        \hline\hline
        YOLOv11 classification backbone with C2PSA & 0.775 \\
        YOLOv11 classification backbone with C2PCA & 0.793 \\    
        \hline
    \end{tabularx}
\end{table}

\subsubsection{Effect of adding Feature Extraction and Aggregation (FEA) modules}
We progressively insert the proposed FEA modules at three hierarchical positions starting from deeper layers within the backbone, as highlighted in the architecture diagram, Fig \ref{aggrnet}.

\begin{itemize}
  \item YOLOv11 classification backbone + C2PCA + FEA@1: FEA module at Position 1.
  \item YOLOv11 YOLOv11 classification backbone + C2PCA + FEA@1,2: FEA modules at Positions 1 and 2.
  \item YOLOv11 YOLOv11 classification backbone + C2PCA + FEA@1,2,3: FEA modules at Positions 1, 2, and 3.
\end{itemize}


This incremental approach isolates the individual effects of each FEA block. As shown in Table \ref{sppf_abalation}, the performance improves steadily with each additional FEA module. 


\subsubsection{Effect of adding Spatial Pyramid Pooling Fast (SPPF) layer}

Finally, we examine the impact of integrating the SPPF block in the proposed AGGRNet architecture. With SPPF added, as shown in Table \ref{sppf_abalation}, AGGRNet achieves the highest accuracy of 0.81, establishing the benefit of adding SPPF block alongside hierarchical feature aggregation.


\begin{table}
    \centering
    \caption{Effect of adding FEA module and SPPF block.}
    \label{sppf_abalation}
    \begin{tabularx}{\linewidth}{X|c} 
        \hline
        \textbf{Architecture} & \textbf{Accuracy} \\ 
        \hline\hline
        YOLOv11 + C2PCA + FEA@1 & 0.791 \\
        YOLOv11 + C2PCA + FEA@1,2 & 0.793 \\
        YOLOv11 + C2PCA + FEA@1,2,3 & 0.807 \\ 
        YOLOv11 + C2PCA + FEA@1,2,3 & 0.807 \\
        \rowcolor[gray]{0.9}\textbf{AGGRNet/Ours (YOLOv11 + C2PCA + FEA@1,2,3 + SPPF)} & \textbf{0.810} \\    
        \hline
    \end{tabularx}
\end{table}


\section{Conclusion}
\label{sec:conclusion}
In this paper, we propose a novel framework, AGGRNet, with the capability to capture both informative and non-informative features using the proposed Feature Extraction and Aggregation (FEA) module for effectively classifying disease subtypes and severity. We validated the proposed AGGRNet framework through extensive experiments on multiple datasets, while improving the medical image classification performance over the state-of-the-art (SOTA) models.  The detailed ablation studies highlight the importance of each proposed module to improve the performance. This approach enables clinicians to diagnose and treat patients more accurately, reducing reliance on manual scoring methods that are often prone to subjective ambiguity. 

{
    \small
    \bibliographystyle{ieeenat_fullname}
    \bibliography{main}

\begin{thebibliography}{33}
\providecommand{\natexlab}[1]{#1}
\providecommand{\url}[1]{\texttt{#1}}
\expandafter\ifx\csname urlstyle\endcsname\relax
  \providecommand{\doi}[1]{doi: #1}\else
  \providecommand{\doi}{doi: \begingroup \urlstyle{rm}\Url}\fi

\bibitem[Albuquerque et~al.(2021)Albuquerque, Cruz, and Cardoso]{albuquerque2021ordinal}
Tom{\'e} Albuquerque, Ricardo Cruz, and Jaime~S Cardoso.
\newblock Ordinal losses for classification of cervical cancer risk.
\newblock \emph{PeerJ Computer Science}, 7:\penalty0 e457, 2021.

\bibitem[Codella et~al.(2018)Codella, Rotemberg, Tschandl, Celebi, Dusza, Gutman, Helba, Kalloo, Liopyris, Marchetti, Kittler, and Halpern]{ISIC1}
Noel C.~F. Codella, Veronica Rotemberg, Philipp Tschandl, M.~Emre Celebi, Stephen~W. Dusza, David Gutman, Brian Helba, Aadi Kalloo, Konstantinos Liopyris, Michael Marchetti, Harald Kittler, and Allan Halpern.
\newblock Skin lesion analysis toward melanoma detection 2018: A challenge hosted by the international skin imaging collaboration (isic), 2018.
\newblock arXiv preprint arXiv:1902.03368.

\bibitem[Dosovitskiy et~al.(2020)Dosovitskiy, Beyer, Kolesnikov, Weissenborn, Zhai, Unterthiner, Dehghani, Minderer, Heigold, Gelly, Uszkoreit, and Houlsby]{ViT}
Alexey Dosovitskiy, Lucas Beyer, Alexander Kolesnikov, Dirk Weissenborn, Xiaohua Zhai, Thomas Unterthiner, Mostafa Dehghani, Matthias Minderer, Georg Heigold, Sylvain Gelly, Jakob Uszkoreit, and Neil Houlsby.
\newblock An image is worth 16x16 words: Transformers for image recognition at scale.
\newblock \emph{ArXiv}, abs/2010.11929, 2020.

\bibitem[Hashash et~al.(2024)Hashash, Yu~Ci~Ng, Farraye, Wang, Colucci, Baxi, Muneer, Reddan, Shingru, and Melmed]{hashash2024inter}
Jana~G Hashash, Faye Yu~Ci~Ng, Francis~A Farraye, Yeli Wang, Daniel~R Colucci, Shrujal Baxi, Sadaf Muneer, Mitchell Reddan, Pratik Shingru, and Gil~Y Melmed.
\newblock Inter-and intraobserver variability on endoscopic scoring systems in crohn’s disease and ulcerative colitis: a systematic review and meta-analysis.
\newblock \emph{Inflammatory Bowel Diseases}, 30\penalty0 (11):\penalty0 2217--2226, 2024.

\bibitem[He et~al.(2016{\natexlab{a}})He, Zhang, Ren, and Sun]{deep-residual-learning}
Kaiming He, Xiangyu Zhang, Shaoqing Ren, and Jian Sun.
\newblock Deep residual learning for image recognition.
\newblock In \emph{2016 IEEE Conference on Computer Vision and Pattern Recognition (CVPR)}, pages 770--778, 2016{\natexlab{a}}.

\bibitem[He et~al.(2016{\natexlab{b}})He, Zhang, Ren, and Sun]{he2016deep}
Kaiming He, Xiangyu Zhang, Shaoqing Ren, and Jian Sun.
\newblock Deep residual learning for image recognition.
\newblock In \emph{Proceedings of the IEEE conference on computer vision and pattern recognition}, pages 770--778, 2016{\natexlab{b}}.

\bibitem[Howard et~al.(2019)Howard, Sandler, Chu, Chen, Chen, Tan, Wang, Zhu, Pang, Vasudevan, et~al.]{howard2019searching}
Andrew Howard, Mark Sandler, Grace Chu, Liang-Chieh Chen, Bo Chen, Mingxing Tan, Weijun Wang, Yukun Zhu, Ruoming Pang, Vijay Vasudevan, et~al.
\newblock Searching for mobilenetv3.
\newblock In \emph{Proceedings of the IEEE/CVF international conference on computer vision}, pages 1314--1324, 2019.

\bibitem[Huang et~al.(2017)Huang, Liu, Van Der~Maaten, and Weinberger]{huang2017densely}
Gao Huang, Zhuang Liu, Laurens Van Der~Maaten, and Kilian~Q Weinberger.
\newblock Densely connected convolutional networks.
\newblock In \emph{Proceedings of the IEEE conference on computer vision and pattern recognition}, pages 4700--4708, 2017.

\bibitem[Huo et~al.(2022)Huo, Sun, Tian, Wang, Yu, Long, Zhang, and Li]{HiFuse}
Xiangzuo Huo, Gang Sun, Sheng Tian, Yan Wang, Long Yu, Jun Long, Wendong Zhang, and Aolun Li.
\newblock Hifuse: Hierarchical multi-scale feature fusion network for medical image classification.
\newblock \emph{Biomed. Signal Process. Control.}, 87:\penalty0 105534, 2022.

\bibitem[Huo et~al.(2024)Huo, Sun, Tian, Wang, Yu, Long, Zhang, and Li]{huo2024hifuse}
Xiangzuo Huo, Gang Sun, Shengwei Tian, Yan Wang, Long Yu, Jun Long, Wendong Zhang, and Aolun Li.
\newblock Hifuse: Hierarchical multi-scale feature fusion network for medical image classification.
\newblock \emph{Biomedical Signal Processing and Control}, 87:\penalty0 105534, 2024.

\bibitem[Khanam and Hussain(2024)]{Khanam2024YOLOv11AO}
Rahima Khanam and Muhammad Hussain.
\newblock Yolov11: An overview of the key architectural enhancements, 2024.

\bibitem[Li et~al.(2023)Li, Wang, Zhang, Gao, Song, Liu, Li, and Qiao]{Uniformer}
Kunchang Li, Yali Wang, Junhao Zhang, Peng Gao, Guanglu Song, Yu Liu, Hongsheng Li, and Yu Qiao.
\newblock Uniformer: Unifying convolution and self-attention for visual recognition.
\newblock \emph{IEEE transactions on pattern analysis and machine intelligence}, PP, 2023.

\bibitem[Lin et~al.(2022)Lin, Cheng, Wu, and Shen]{lin2022cat}
Hezheng Lin, Xing Cheng, Xiangyu Wu, and Dong Shen.
\newblock Cat: Cross attention in vision transformer.
\newblock In \emph{2022 IEEE international conference on multimedia and expo (ICME)}, pages 1--6. IEEE, 2022.

\bibitem[Liu et~al.(2021)Liu, Lin, Cao, Hu, Wei, Zhang, Lin, and Guo]{Swin}
Ze Liu, Yutong Lin, Yue Cao, Han Hu, Yixuan Wei, Zheng Zhang, Stephen Lin, and Baining Guo.
\newblock Swin transformer: Hierarchical vision transformer using shifted windows.
\newblock In \emph{2021 IEEE/CVF International Conference on Computer Vision (ICCV)}, pages 9992--10002, 2021.

\bibitem[Liu et~al.(2022)Liu, Mao, Wu, Feichtenhofer, Darrell, and Xie]{ConvNext}
Zhuang Liu, Hanzi Mao, Chao-Yuan Wu, Christoph Feichtenhofer, Trevor Darrell, and Saining Xie.
\newblock A convnet for the 2020s.
\newblock In \emph{2022 IEEE/CVF Conference on Computer Vision and Pattern Recognition (CVPR)}, pages 11966--11976, 2022.

\bibitem[Min et~al.(2022)Min, Zhao, Luo, and Cho]{PerViT}
Juhong Min, Yucheng Zhao, Chong Luo, and Minsu Cho.
\newblock Peripheral vision transformer, 2022.

\bibitem[Peng et~al.(2021)Peng, Huang, Gu, Xie, Wang, Jiao, and Ye]{Conformer}
Zhiliang Peng, Wei Huang, Shanzhi Gu, Lingxi Xie, Yaowei Wang, Jianbin Jiao, and Qixiang Ye.
\newblock Conformer: Local features coupling global representations for visual recognition.
\newblock \emph{2021 IEEE/CVF International Conference on Computer Vision (ICCV)}, pages 357--366, 2021.

\bibitem[Pogorelov et~al.(2017)Pogorelov, Randel, Griwodz, Eskeland, de~Lange, Johansen, Spampinato, Dang-Nguyen, Lux, Schmidt, Riegler, and Halvorsen]{Kvasir}
Konstantin Pogorelov, Kristin~Ranheim Randel, Carsten Griwodz, Sigrun~Losada Eskeland, Thomas de Lange, Dag Johansen, Concetto Spampinato, Duc-Tien Dang-Nguyen, Mathias Lux, Peter~Thelin Schmidt, Michael Riegler, and P{\aa}l Halvorsen.
\newblock Kvasir: A multi-class image dataset for computer aided gastrointestinal disease detection.
\newblock In \emph{Proceedings of the 8th ACM on Multimedia Systems Conference}, pages 164--169, New York, NY, USA, 2017. ACM.

\bibitem[Polat et~al.(2022{\natexlab{a}})Polat, Ergenc, Kani, Alahdab, Atug, and Temizel]{Polat22}
Gorkem Polat, Ilkay Ergenc, Haluk~Tarik Kani, Yesim~Ozen Alahdab, Ozlen Atug, and Alptekin Temizel.
\newblock Class distance weighted cross-entropy loss for ulcerative colitis severity estimation, 2022{\natexlab{a}}.

\bibitem[Polat et~al.(2022{\natexlab{b}})Polat, Kani, Ergenc, Alahdab, Temizel, and Atug]{LIMUC}
Görkem Polat, Haluk~Tarik Kani, Ilkay Ergenc, Yesim~Ozen Alahdab, Alptekin Temizel, and Ozlen Atug.
\newblock Labeled images for ulcerative colitis (limuc) dataset, 2022{\natexlab{b}}.
\newblock Zenodo Dataset, version 1, DOI: 10.5281/zenodo.5827695.

\bibitem[Sharara et~al.(2022)Sharara, Malaeb, Lenfant, and Ferrante]{sharara2022assessment}
Ala~I Sharara, Maher Malaeb, Matthias Lenfant, and Marc Ferrante.
\newblock Assessment of endoscopic disease activity in ulcerative colitis: is simplicity the ultimate sophistication?
\newblock \emph{Inflammatory Intestinal Diseases}, 7\penalty0 (1):\penalty0 7--12, 2022.

\bibitem[Shi et~al.(2023)Shi, Cao, and Raschka]{shi2023deep}
Xintong Shi, Wenzhi Cao, and Sebastian Raschka.
\newblock Deep neural networks for rank-consistent ordinal regression based on conditional probabilities.
\newblock \emph{Pattern Analysis and Applications}, 26\penalty0 (3):\penalty0 941--955, 2023.

\bibitem[Shiku et~al.(2025)Shiku, Nishimura, Suehiro, Tanaka, and Bise]{OrdinalMultiInstance}
Kaito Shiku, Kazuya Nishimura, Daiki Suehiro, Kiyohito Tanaka, and Ryoma Bise.
\newblock { Ordinal Multiple-instance Learning for Ulcerative Colitis Severity Estimation with Selective Aggregated Transformer }.
\newblock In \emph{2025 IEEE/CVF Winter Conference on Applications of Computer Vision (WACV)}, pages 4290--4299, Los Alamitos, CA, USA, 2025. IEEE Computer Society.

\bibitem[Simonyan and Zisserman(2014)]{VGG}
Karen Simonyan and Andrew Zisserman.
\newblock Very deep convolutional networks for large-scale image recognition.
\newblock \emph{arXiv 1409.1556}, 2014.

\bibitem[Szegedy et~al.(2015)Szegedy, Liu, Jia, Sermanet, Reed, Anguelov, Erhan, Vanhoucke, and Rabinovich]{szegedy2015going}
Christian Szegedy, Wei Liu, Yangqing Jia, Pierre Sermanet, Scott Reed, Dragomir Anguelov, Dumitru Erhan, Vincent Vanhoucke, and Andrew Rabinovich.
\newblock Going deeper with convolutions.
\newblock In \emph{Proceedings of the IEEE conference on computer vision and pattern recognition}, pages 1--9, 2015.

\bibitem[Tolstikhin et~al.(2021)Tolstikhin, Houlsby, Kolesnikov, Beyer, Zhai, Unterthiner, Yung, Steiner, Keysers, Uszkoreit, Lucic, and Dosovitskiy]{mlp-mixer}
Ilya~O. Tolstikhin, Neil Houlsby, Alexander Kolesnikov, Lucas Beyer, Xiaohua Zhai, Thomas Unterthiner, Jessica Yung, Andreas Steiner, Daniel Keysers, Jakob Uszkoreit, Mario Lucic, and Alexey Dosovitskiy.
\newblock Mlp-mixer: An all-mlp architecture for vision.
\newblock \emph{CoRR}, abs/2105.01601, 2021.

\bibitem[Touvron et~al.(2021)Touvron, Cord, Douze, Massa, Sablayrolles, and Jegou]{DeiT-base}
Hugo Touvron, Matthieu Cord, Matthijs Douze, Francisco Massa, Alexandre Sablayrolles, and Herve Jegou.
\newblock Training data-efficient image transformers \&; distillation through attention.
\newblock In \emph{Proceedings of the 38th International Conference on Machine Learning}, pages 10347--10357. PMLR, 2021.

\bibitem[Tschandl et~al.(2018)Tschandl, Rosendahl, and Kittler]{ISIC2}
Philipp Tschandl, Cliff Rosendahl, and Harald Kittler.
\newblock The ham10000 dataset, a large collection of multi-source dermatoscopic images of common pigmented skin lesions.
\newblock \emph{Scientific Data}, 5:\penalty0 180161, 2018.

\bibitem[Woo et~al.(2018)Woo, Park, Lee, and Kweon]{woo2018cbam}
Sanghyun Woo, Jongchan Park, Joon-Young Lee, and In~So Kweon.
\newblock Cbam: Convolutional block attention module.
\newblock In \emph{Proceedings of the European conference on computer vision (ECCV)}, pages 3--19, 2018.

\bibitem[Yang et~al.(2022)Yang, Li, Dai, and Gao]{yang2022focal}
Jianwei Yang, Chunyuan Li, Xiyang Dai, and Jianfeng Gao.
\newblock Focal modulation networks.
\newblock In \emph{Advances in Neural Information Processing Systems}, 2022.

\bibitem[Yang et~al.(2023)Yang, Shi, Wei, Liu, Zhao, Ke, Pfister, and Ni]{Medmnist}
Jiancheng Yang, Rui Shi, Donglai Wei, Zequan Liu, Lin Zhao, Bilian Ke, Hanspeter Pfister, and Bingbing Ni.
\newblock Medmnist v2 - a large-scale lightweight benchmark for 2d and 3d biomedical image classification.
\newblock \emph{Scientific Data}, 10\penalty0 (1), 2023.

\bibitem[Yuan et~al.(2021)Yuan, Chen, Wang, Yu, Shi, Jiang, Tay, Feng, and Yan]{T2T-Vit}
Li Yuan, Yunpeng Chen, Tao Wang, Weihao Yu, Yujun Shi, Zihang Jiang, Francis Tay, Jiashi Feng, and Shuicheng Yan.
\newblock Tokens-to-token vit: Training vision transformers from scratch on imagenet.
\newblock pages 538--547, 2021.

\bibitem[Zhu et~al.(2023)Zhu, Wang, Ke, Zhang, and Lau]{biformer}
Lei Zhu, Xinjiang Wang, Zhanghan Ke, Wayne Zhang, and Rynson Lau.
\newblock Biformer: Vision transformer with bi-level routing attention, 2023.

\end{thebibliography}
}

\end{document}